\documentclass[10pt,twocolumn,letterpaper]{article}

\usepackage{iccv}
\usepackage{times}
\usepackage{epsfig}
\usepackage{graphicx}
\usepackage{amsmath}
\usepackage{amssymb}
\usepackage{url}
\newcommand\numberthis{\addtocounter{equation}{1}\tag{\theequation}}

\usepackage[pagebackref=true,breaklinks=true,letterpaper=true,colorlinks,bookmarks=false]{hyperref}

\iccvfinalcopy % *** Uncomment this line for the final submission

\def\iccvPaperID{****} % *** Enter the ICCV Paper ID here

\ificcvfinal\fi

\begin{document}

%%%%%%%%% TITLE
\title{BERT for Large-scale Video Segment Classification with Test-time Augmentation}

\author{Tianqi Liu\thanks{These two authors contributed equally. The code is publicly available at \url{https://github.com/hughshaoqz/3rd-Youtube8M-TM}.}\\
{\tt\small tianqi.terence.liu@gmail.com}
\and
Qizhan Shao\footnotemark[1]\\
{\tt\small hugh.shao.qz@gmail.com}
}

\maketitle
\ificcvfinal\thispagestyle{empty}\fi

%%%%%%%%% ABSTRACT
\begin{abstract}
   This paper presents our approach to the third YouTube-8M video understanding competition that challenges participants to localize video-level labels at scale to the precise time in the video where the label actually occurs. Our model is an ensemble of frame-level models such as Gated NetVLAD and NeXtVLAD and various BERT models with test-time augmentation. We explore multiple ways to aggregate BERT outputs as video representation and various ways to combine visual and audio information. We propose test-time augmentation as shifting video frames to one left or right unit, which adds variety to the predictions and empirically shows improvement in evaluation metrics. We first pre-train the model on the 4M training video-level data, and then fine-tune the model on 237K annotated video segment-level data. We achieve MAP@100K 0.7871 on private testing video segment data, which is ranked 9th over 283 teams.
\end{abstract}

%%%%%%%%% BODY TEXT
\section{Introduction}
Videos are very popular and important contents on Internet powered by the prevalence of digital cameras and smart phones. Video understanding is one of the major challenges in computer vision and has various applications such as recommendation, searching, smart homes, autonomous driving, and sports video analysis. 

Large scale datasets such as the Sports-1M dataset~\cite{karpathy2014large} and the ActivityNet dataset~\cite{caba2015activitynet} enable researchers to evaluate and compare among methods on video classification of sports and human activities. In a larger scale and more comprehensive way, the YouTube-8M dataset~\cite{abu2016youtube} consists of 6 million of YouTube video IDs with high-quality annotations of 3,800+ visual entities. Each video is decoded at 1 frame-per-second up to the first 360 seconds, after which features are extracted via pre-trained model. PCA and quantization are further applied to reduce dimensions and data size. Visual features of 1024 dimensions and audio features of 128 dimensions are extracted on each frame as input for downstream classifiers. Several existing competitions have been held to build both \href{https://www.kaggle.com/c/youtube8m}{unconstrained} and \href{https://www.kaggle.com/c/youtube8m-2018}{constrained} models for video-level annotations. 

In most web searches, video retrieval and ranking can be achieved by matching query terms to video-level features. However, many of them miss temporal localization to important moments within the video. Temporal localization can enable search within video, video highlight extraction, safer video content, and many others. Motivated by this, the YouTube-8M Segments dataset extends the YouTube-8M dataset with human-verified segment annotations and enables temporal localization of the entities in the videos. The YouTube-8M Segments dataset collects human-verified 237K segments on 1000 classes from the validation set of the YouTube-8M dataset. The dataset enables classifier to predict at 5 frames segment-level granularity. 

Based on the YouTube-8M Segments dataset, the 3rd YouTube-8m video understanding challenge challenges participants to build machine learning models to localize video-level labels to the precise time in the video where the label actually appears. Submissions are evaluated based on the Mean Average Precision @ K (MAP@K), where K = 100,000.
\begin{equation}
    \text{MAP@100,000} = \frac{1}{C}\sum_{c=1}^C \frac{\sum_{k=1}^KP(k)\times rel(k)}{N_c},
\end{equation}
where $C$ is the number of classes, $P(k)$ is the precision at cutoff $k$, $K$ is the number of segments predicted per class, $rel(k)$ is an indicator function equaling 1 if the item at rank $k$ is a relevant (correct) class, or zero otherwise, and $N_c$ is the number of positively-labeled segments for the each class.

Common methods for video analysis involve temporal aggregation of visual and audio features by learnable pooling methods such as generalized vector of locally aggregated descriptors (NetVLAD)~\cite{arandjelovic2016netvlad}, deep bag of frames (DBoF)~\cite{abu2016youtube}, convolutional neural network (CNN)~\cite{karpathy2014large}, gated recurrent unit (GRU)~\cite{cho2014properties}, and long short-term memory (LSTM)~\cite{hochreiter1997long}. More recently, Gated NetVLAD~\cite{miech2017learnable}, NeXtVLAD~\cite{Lin_2019}, and VideoBERT~\cite{sun2019videobert} show advantage and popularity in different applications. Among the aforementioned models, recurrent models (LSTM and GRU) capture time dependencies as temporal aggregration of variable-length sequences, while their training requires large amount of data and can be sub-optimal for long video sequences~\cite{miech2017learnable}. Orderless aggregation approachs (NetVLAD, DBoF, NeXtVLAD) show better performance on video classification in previous studies~\cite{miech2017learnable, abu2016youtube, Lin_2019}. In the first YouTube-8M video understanding challenge~\cite{miech2017learnable, wang2017monkeytyping, li2017temporal, chen2017aggregating, skalic2017deep}, Gated NetVLAD, LSTM, and GRU followed by logistic regression~\cite{kleinbaum2002logistic} or mixture of experts (MoE)~\cite{jordan1994hierarchical} are shown to be of the best performance. The second YouTube-8M video understanding~\cite{skalic2018building, ostyakov2018label, Lin_2019, tang2018non, kim2018temporal, liu2018constrained} demonstrates the advantage of knowledge distillation~\cite{hinton2015distilling}, weights quantization~\cite{han2015deep}, dimensional reduction~\cite{Lin_2019, xie2017aggregated, liu2017characterizing}, conditional inference~\cite{kim2018temporal}, and float16 inference~\cite{liu2018constrained} in learning efficient video classification models.

For video segment classification and temporal localization, transfer learning~\cite{pan2009survey} leveraging existing dataset on video-level labels can serve as a warm-start for training segment-level models. Several directions can also be relevant and helpful for improving the models. For example, segment-level classification can be treated as multiple-instance learning (MIL)~\cite{zhou2006multi, ilse2018attention} in two ways: one way is to treat video-level data (generally around 300 frames) as multiple instances of video segment; the other way is to treat each frame as one instance so that the entity mention in each segment is true if and only if at least one frame contains the information. Typical MIL is implemented by mean, max, or attention-based pooling~\cite{ilse2018attention}. Another direction is semi-supervised learning approach~\cite{ando2005framework, xie2019unsupervised}. Since segment-level data is quite limited, semi-supervised learning approach can potentially add value by leveraging large amount of unlabeled data. Effective data augmentation approach can also improve the model robustness and help with semi-supervised learning.

In this work, we make the following three contributions. First we explore the usage of transfer learning leveraging noisy video-level data on segment-level classification and compare different methods such as NetVLAD, NeXtVLAD, and BERT. Second we investigate and demonstrate the usage of BERT model on video classification. Third we propose a simple yet useful test-time augmentation approach that can add variation to prediction and boost the evaluation score.

\section{Related work}
\paragraph{BERT}
For recent progress in the natural language processing (NLP) community, large-scale language models such as embeddings from language models (ELMO)~\cite{peters2018deep}, bidirectional encoder representations from transformers (BERT)~\cite{devlin2018bert}, and XLNet~\cite{yang2019xlnet} have shown state-of-the-art results of various NLP tasks, both at word level such as POS tagging and sentence level such as sentiment analysis. Motivated by the good performance of BERT on sequence modeling, we build a BERT model on frame-level features to enforce attention mechanism and enable the long term dependency modeling for video-level and segment-level classification. 

\paragraph{Frame-level models}
Video features are typically extracted from individual frames by deep convolutional neural networks. There are two ways to aggregate them: Order and orderless ways. 

Order and time information can be modeled by applying recurrent neural networks such as LSTM~\cite{hochreiter1997long} and GRU~\cite{cho2014properties} on top of extracted frame-level features~\cite{donahue2015long, fernando2015modeling, ibrahim2016hierarchical}. Hieracrchical spatio-temporal convolution architectures~\cite{tran2015learning, baccouche2011sequential, carreira2017quo, feichtenhofer2017spatiotemporal} can extract and aggregate temporal features at the same time. 

Orderless way captures only the distribution of features in the video. The simplest way is the average or maximum pooling of video features~\cite{wang2016temporal}. Advanced methods include bag-of-visual-words~\cite{csurka2004visual}, DBoF~\cite{abu2016youtube}, Gated NetVLAD~\cite{miech2017learnable}, and NeXtVLAD~\cite{Lin_2019}. In our work, we use Gated NetVLAD and NeXtVLAD as one of our models for their good performance in the previous competitions~\cite{miech2017learnable, Lin_2019}.

\paragraph{Video data augmentation}
Since video is a sequence of frames and audios. standard image augmentation approach~\cite{perez2017effectiveness, cubuk2018autoaugment} such as flip, rotation, and zooming in, can be applied for video data augmentation. However in our case, the frame-level visual and audio features are extracted by pre-trained network, we cannot augment on raw data. Random sampling with replacement and random sub-sequence sampling are shown to be effective in previous competitions~\cite{miech2017learnable, Lin_2019, wang2017monkeytyping}. In our work, we apply the same sampling approach for data augmentation for training. In inference, we propose an easy augmentation approach by shifting segment-level data and ensembling them.

\section{BERT for video classification}
\begin{figure}[t]
\begin{center}
  \includegraphics[width=0.8\linewidth]{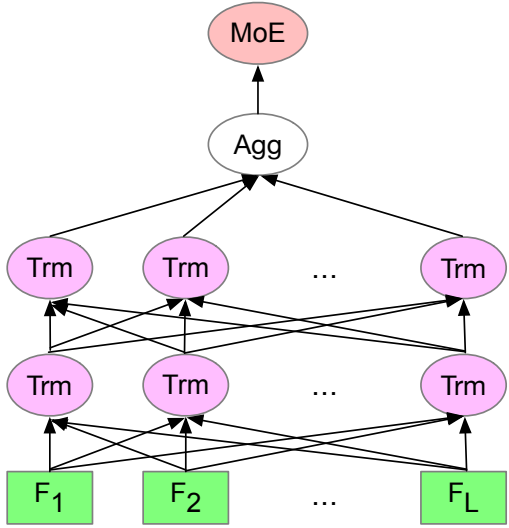}
\end{center}
   \caption{Our proposed BERT model for video segment classification. $F_1, ..., F_L$ represents input frame-level features, Trm is short for Transformer, Agg is short for aggregation function, and MoE is short for mixture of experts classifier. Agg can be one of ``first'', ``mean'', ``attention''. ``First'' means using first token of BERT output layer. ``Mean'' means averaging all the outputs from the last layer. ``attention'' means learnable weighted averaging of all the outputs from last layer of BERT.}
\label{fig:bert}
\end{figure}

\paragraph{Our BERT for video classification}
Figure \ref{fig:bert} illustrates our proposed BERT model for video classification. The model takes pre-processed frame-level visual and audio features as input, applies BERT, and aggregates outputs from BERT into a video-level representation. Finally, we use Mixture of Expert (MoE) network as the last layer of the classifier. Several aggregation functions have been used: ``first'', ``mean'', and ``attention''. ``First'' means we use the first output token from BERT last layer as video representation. ``Mean'' means taking the average over tokens from BERT last layer. ``Attention'' means applying an attention function to compute weighted average over tokens from BERT last layer.

\paragraph{BERT review}
The BERT model~\cite{devlin2018bert} with hidden dimension $H$ uses a list of discrete $L$ tokens $t={t_1, . . . , t_L}$ as inputs. There are mainly two components in the model: the encoder $f_{enc}(t_l)$, which is an embedding lookup table that maps the token $t_l = i$ to a feature vector $f_i\in\mathbb{R}^{H}$; and a context-based predictor $g_{pred}(T_{\setminus l})$ ($T_{\setminus l}$ means tokens except $t_l$), which is a multi-layer multi-head transformer network~\cite{vaswani2017attention} that takes $L\times H$ feature matrix as inputs, returns the same size matrix, and then outputs $\hat{f}_l\in\mathbb{R}^{H}$ as the prediction of $f_i$. In order to make this $t_l$ token's prediction, the ground-truth feature at the $l$-th row of the input matrix is masked out, and the $l$-th row of the output matrix is used as $\hat{f}_l$. BERT can mask more than one token in each sentence.

BERT requires a fixed discrete vocabulary to compute final predictions, but for image and videos, the inputs are continuous. So we can not use BERT directly. One way is to apply VideoBERT~\cite{sun2019videobert}, which uses the softmax version of noise contrastive estimation to calculate prediction probabilities. However, since we already have video-level labels for pre-training, we can simply pass the frame-level visual and audio features into BERT, and utilize the power of multi-head self-attention to learn the video-level representation.

The Transformer~\cite{vaswani2017attention} uses multi-head attention mechanism which consists of several scaled dot-product attention layers running in parallel. The input of each layer consists of a set of queries of dimension $d_k$, packed together into a matrix ${Q}$, and a set of keys and values of dimension $d_k$, $d_v$, packed together into matrices ${K}$ and ${V}$. The output of each layer is calculated as:
\begin{equation}
    \text{Attention}({Q},{K},{V}) = \text{softmax}(\frac{{Q}{K}^T}{\sqrt{d_k}}){V}
\end{equation}
We can see each layer as a head, and the multi-head mechanism first linearly projects the queries, keys, and values to calculate each head, and then concatenates all the heads and once again do the linear projection as the following way to get final results:
\begin{equation}
    \text{MultiHead}({Q},{K},{V}) = \text{Concat}(\text{head}_1, ..., \text{head}_h){W}^O,
\end{equation}
where $\text{head}_i$ = $\text{Attention}({Q}{W}^Q_i$, ${K}{W}^K_i$, ${V}{W}^V_i$) \\
and $h$ is number of heads, ${W}^Q_i\in\mathbb{R}^{d_{m}\times d_k}$, ${W}^K_i\in\mathbb{R}^{d_{m}\times d_k}$, ${W}^V_i\in\mathbb{R}^{d_{m}\times d_v}$, ${W}^O_i\in\mathbb{R}^{hd_{v}\times d_m}$ are projection parameter matrices with $d_k=d_v=d_m/h$, where $d_m$ is the dimension of the original feature vector of each token.

In our case, all the ${Q}$, ${K}$, ${V}$ are the same, which are frame-level visual and audio feature vectors.

\paragraph{Cross-modal Learning}
Video has multi-modal nature since most videos have synchronized frame-level visual and audio signals. We can make these two modalities supervise each other to learn more robust self-supervised video representation. Similar to the approaches Sun \etal~\cite{sun2019videobert, 2019arXiv190605743S} used, we first apply BERT to learn frame-level visual and audio representation separately, and then concatenate both representations followed by another layer of Transformer. Let frame-level visual feature be ${F_{visual}}$, frame-level audio feature be ${F_{audio}}$, then the cross-modal method is:
\begin{equation}
    {B} = T_{cross}([T_{visual}({F_{visual}});T_{audio}({F_{audio}})]),
\end{equation}
where $T_{visual}$ and $T_{audio}$ are Transformers applied to visual and audio signals, repectively. And $T_{cross}$ is a Transformer applied on top of concatenated outputs of transformed visual and audio signals.

On the other hand, we also have models that first concatenate visual and audio signals in each frame and feed into transformer directly.

\paragraph{Video-level representation}
We apply some aggregation function to obtain vector representation of video. Let ${b}_i\in \mathbb{R}^{d_m}$ be the output of last layer by BERT, the representation of video $v$ is an aggregation of $b_i$s. There are three ways of aggregation: using the first frame, taking average, and taking learnable weighted average.

We learn the attention weights for weighted average as
\begin{equation}
    a_l = \frac{e^{w^T{b}_l}}{\sum_{j=1}^Le^{w^T{b}_l}}.
\end{equation}
And video representation $v$ is 
\begin{equation}
    {v} = \sum_{l=1}^La_l{b}_l,
\end{equation}

\section{Frame-level models}
\paragraph{Gated NetVLAD}
Consider a video with $I$ frames and $J$-dimensional frame-level features $\{x_i\}_{i=1}^I$ extracted by a pre-trained CNN. $K$ clustered NetVLAD first encodes each feature into $J\times K$ dimension using the following equation:
\begin{equation}
    v_{ijk} = \alpha_k(x_i)(x_{ij} - c_{kj}), i\in [I], j\in [J], k\in [K],
\end{equation}
where $[n]$ stands for $\{1,..,n\}$, $c_k$ is the learnable $J$-dimensional anchor point of cluster $k$, and $\alpha_k(x_i)$ is a soft assignment function of $x_i$ to cluster $k$. 
\begin{equation}
    \alpha_k(x_i) = \frac{e^{w_k^Tx_i + b_k}}{\sum_{l=1}^Ke^{w_l^Tx_i + b_l}},
\end{equation}
where $\{w_k\}_{k=1}^K$ and ${b_k}_{k=1}^K$ are learnable parameters.

Second, a video-level descriptor $y$ is obtained by aggregating all the encoded frame-level features,
\begin{equation}
    y_{jk} = \sum_{i=1}^Iv_{ijk},
\end{equation}
followed by normalization to unit vector across dimension $j$. 

Third, the constructed video-level descriptor $y$ is reduced to an $H$-dimensional hidden vector via a fully-connected layer. And context gating (CG) applies to this video hidden representation as 
\begin{equation}
    z = \sigma(Wy+b)\circ y,
\end{equation}
where $y\in\mathbb{R}^{JK}$ is the input feature vector, $\sigma$ is the element-wise sigmoid activation and 
$\circ$ is the element-wise multiplication. $W\in\mathbb{R}^{JK\times JK}$ and $b\in\mathbb{R}^{JK}$ are trainable parameters. 

Finally, context gated vector $z$ is fed into MoE classifier followed by another CG to compute the logits.

NetVLAD can provide an aggregated vector representation of the video and CG can add non-linear interactions among input vector and recalibrate the strengths of different dimensions of input representation.

\paragraph{NeXtVLAD}
In the NeXtVLAD~\cite{Lin_2019} aggregation network, the frame-level feature $x_i$ is first expanded to $\dot{x}_i$ with a dimension of $\lambda J$ through a linear fully-connected layer, where $\lambda$ is set to be $2$ in our experiments. Then $\dot{x}$ with shape $(I, \lambda J)$ is reshaped to $\tilde{x}$ with shape $(I, G, \lambda J / G)$ with $G$ as the size of groups. Each of $\tilde{x}_i^g\in\mathbb{R}^{\lambda J/G}$ is transformed to NetVLAD representation in the following way:
\begin{align*}
    &v_{ijk}^g = \alpha_g(\dot{x}_i)\alpha_{gk}(\dot{x}_i)(\tilde{x}^g_{ij}-c_{kj}),\numberthis \label{eqn}\\
   & g\in [G], i\in [I], j\in [\lambda J/G], k \in [K],  
\end{align*}
where $\alpha_g(\dot{x}_i))$ is the attention function over groups with
\begin{equation}
    \alpha_{g}(\dot{x}_i) = \sigma \left(w^T_{g}\dot{x}_i+b_{g}\right),
\end{equation}

and $\alpha_{gk}(\dot{x}_i)$ is the soft assignment of clusters with
\begin{equation}
    \alpha_{gk}(\dot{x}_i) = \frac{e^{w^T_{gk}\dot{x}_i+b_{gk}}}{\sum_{l=1}^K{e^{w^T_{gl}\dot{x}_i+b_{gl}}}}.
\end{equation}
Then the video-level descriptor is obtained by aggregating encoded vectors over time and group:
\begin{equation}
    y_{jk} = \sum_{i,g}v_{ijk}^g,
\end{equation}
after which a $l_2$ normalization is applied across dimension $j$. 

Then, the constructed video-level descriptor $y$ is reduced to an $H$-dimensional hidden vector via a fully-connected layer. Squeeze-and-Excitation Context Gating (SECG)~\cite{hu2018squeeze, Lin_2019} is applied to the hidden representation as an efficient replacement for CG with $16$ times less parameters than CG in our experiment.

We use the same approach as in Lin \etal~\cite{Lin_2019} for knowledge distillation with on-the-fly naive ensemble. We train 3 NeXtVLAD models and the logits of the mixture predictions $z^e$ is a weighted sum of single model logits $\{z^m|m\in[3]\}$:
\begin{equation}
    z^e = \sum_{m=1}^3\alpha_m(\bar{x})z^m,
\end{equation}
where $\bar{x}$ is the frame mean of input features $x$, and
\begin{equation}
    \alpha_m(\bar{x}) = \frac{e^{w_m^T\bar{x} + b_m}}{\sum_l^3{e^{w_l^T\bar{x} + b_l}}}.
\end{equation}
The knowledge of the mixed prediction is distilled to each sub-model by minimizing the KL divergence between the probability predictions:
\begin{equation}
    \mathcal{L}_{kl}^{m,e} = \sum_{c=1}^C p^e(c)\log{\frac{p^e(c)}{p^m(c)}},
\end{equation}
where $C$ is the total number of classes and $p(.)$ is the softmax function for probability prediction:
\begin{equation}
    p^m(c) = \frac{e^{z_c^m/T}}{\sum_{l=1}^Ce^{z_l^m/T}}, p^e(c) = \frac{e^{z_c^e/T}}{\sum_{l=1}^Ce^{z_l^e/T}}, 
\end{equation}
where $T$ is a temperature adjusting relative importance of logits. The final loss of the model is:
\begin{equation}
    \mathcal{L} = \sum_{m=1}^3\mathcal{L}_{bce}^m + \mathcal{L}_{bce}^e + T^2\sum_{m=1}^3\mathcal{L}_{kl}^{m,e},
\end{equation}
where $\mathcal{L}_{bce}$ means the binary cross entropy between the ground truth labels and prediction from the model.

\section{Test-time augmentation}
The longer we watch the video, the better we understand the content. We find that video-level label is easier to predict than segment-level label, which we think is mainly because video segment is much shorter and often contains limited information. By watching several frames before or after the segment could probably add more relevant information to determine the topic and entity. Motivated by this, at inference time we shift the video segment one unit both left and right. And we have two more predictions for free on top of the original one. A simple average ensemble can provide us 1-2\% boost in MAP@100K. The ensemble approach will be introduced in the next section.

\section{Ensemble}
\paragraph{Ensembling approach}
We conduct ensemble directly on the submission files. Suppose we want to ensemble $m$ models with weight $w_1, w_2, ..., w_m$. For one class $c$, suppose model $i$ has the top ranked video segment ID $v_i^{(1)}, ..., v_i^{(n)}$, where $n$ is 100K in our case. Our ensemble approach is as follows:
\begin{enumerate}
    \item For any video segment $v_s$, compute its relevance with class $c$ as 
    \begin{equation}
        \text{Score}(v_s) = \sum_{v_s=v_i^{(j)}}\frac{w_i}{j}.
    \end{equation}
    \item Order the video segments according to the score decreasingly.
\end{enumerate}

\paragraph{Bayesian optimization}
Bayesian optimization~\cite{snoek2012practical} optimizes the function by constructing a posterior distribution which best describes that function. The algorithm will become more confident of finding worth exploring space for the parameters as the number of observations grows. It uses statistical model for objective modeling and applies an acquisition function for the next sample decision~\cite{2018arXiv180702811F}. 

In this competition, we use Bayesian optimization to maximize local MAP@100K, to choose best weights for final ensemble. The local MAP score is calculated based on the 300 out of 3,844 validation partition files that are not used for training.

Suppose we have $m$ models and want to ensemble the models with weights $w_1, ..., w_m$. In order to find best weights, we follow the following steps:
\begin{enumerate}
    \item Initialize their possible regions based on each model's public MAP score.
    \item Calculate local MAP score given the current weights.
    \item Apply Bayesian optimization to tune the weights according to the local MAP score
    \item Repeat 2, 3 until converge.
\end{enumerate}
Once we find the optimal weights, we use the learned weights and apply this ensemble method to test files to get our final submission file. This ensemble approach empirically shows a good performance for MAP optimization.

\section{Training details}
\paragraph{Training and evaluation data split}
We randomly select 300 out of 3,844 validation partition files with both video-level and segment-level labels as local validation dataset. We use the local validation dataset to guide hyper-parameter selection and choose the best checkpoint after fine-tuning.

\paragraph{Pre-training}
For all of our models, we first pre-train the model on total 4M video-level label training partition and almost all validation partition data except the aforementioned 300 holdout files for 200k steps. The batch size is 128 and the initial learning rate is 1e-4. The learning rate is exponentially decreased by a factor of 0.9 every 2M examples. 

\paragraph{Fine-tuning}
After the pre-training, we fine-tune the model on validation partition files with segment-level labels except the aforementioned 300 holdout files for local validation. We fine-tune the model for another 20k steps and pick the checkpoint with the highest MAP on local validation data. 

\paragraph{Computational resources}
Our experiments are done with one NVIDIA GTX1080Ti, one NVIDIA RTX2080Ti, and two GCP accounts with K80 GPUs. Each model takes about 12 hours for training and 10 hours for inference.

\section{Experiments}
\subsection{Transfer learning}
Table \ref{tab: pretrain_finetune} shows around 5\% improvement in MAP by fine-tuning on top of pre-trained model on two of our trained models.

MixNeXtVLAD is an ensemble of 3 NeXtVLAD with logistic regression as video-level classification model. Each NeXtVLAD has 8 groups, 112 clusters, 2048 dimensional hidden size, 2 expansions, 16 gating reduction, and 3 temperature.

BERTMean(L2h12) is a two-layer BERT model with 12 heads in multi-head attention, which averages the last layer of BERT and feeds into MoE classifier.

The numerical results demonstrate the following two facts: 
\begin{itemize}
    \item the 4M data with video-level label contains majority of information applying to predictions on segment-level data.
    \item the 237K segment-level training data can further improve the performance of classifying segment-level entities.
\end{itemize}

\begin{table}[h]
\small
\begin{center}
\begin{tabular}{|l|c|c|}
\hline
Model & Public MAP & Private MAP \\
\hline\hline
MixNeXtVLAD pre-training & 0.7411 & 0.7315 \\
MixNeXtVLAD fine-tuning & 0.7739 & 0.7639 \\
BERT(L2h12) pre-training & 0.7278 & 0.7157 \\
BERT(L2h12) fine-tuning & 0.7614 & 0.7551 \\
\hline
\end{tabular}
\end{center}
\caption{Effect of fine-tuning on segment-level labels from pre-trained video-level labels measured by MAP@100K.}
\label{tab: pretrain_finetune}
\end{table}

\subsection{Test-time augmentation}
Table \ref{tab: tta} illustrates the effect of test-time augmentation on segment-level video classification task. MixNeXtVLAD and BERT(L2h12) achieve 0.94\% and 1.71\% relative improvement in MAP@100K, respectively. We also try different shift such as $[0, 2]$ and $[-2, 1]$ to add diversity to our model portfolio in our final ensemble. Here $[a, b]$ means that we average the inferences obtained by shifting $i$ unit for all integer $i$ with $a\le i\le b$.

\begin{table}[h]
\small
\begin{center}
\begin{tabular}{|l|c|c|}
\hline
Model & Public MAP & Private MAP  \\
\hline\hline
MixNeXtVLAD     & 0.7739 & 0.7639 \\
MixNeXtVLAD TTA[-1, 1] & 0.7809 & 0.7711 \\
BERT(L2h12) & 0.7614 & 0.7551 \\
BERT(L2h12) TTA[-1, 1]  & 0.7729 & 0.7680 \\
\hline
\end{tabular}
\end{center}
\caption{Test-time augmentation on segment class prediction by ensembling three inferences (left shift, right shift, original) can improve MAP@100K.}
\label{tab: tta}
\end{table}

\subsection{Model evaluation}
Our final model is an ensemble of 9 models: 1 GatedNetVLAD, 2 MixNeXtVLAD, and 6 BERT models with various layers, number of heads, and aggregation approachs. Our best single model on private dataset is BERTMean(L2h12) TTA[-2,1] with MAP@100K 0.7725. The model is a two-layer BERT model with 12 heads in multi-head attention, which averages the last layer of BERT and feeds into MoE classifier. For TTA, the model ensembles inferences by shifting -2, -1, 0, 1 units of frames. MixNeXtVLAD also shows its good performance on segment-level video classification, probably because of the ensemble and good generalization ability by dimensional reduction.

We use various TTA shift units to increase the variability in our model portfolio. We believe that careful choice of shifting units can further improve the ensemble score. Our single models' weights and MAP@100Ks on public and private dataset are summarized in Table \ref{tab: evaluation}.

\begin{table*}[h]
\begin{center}
\begin{tabular}{|l|c|c|c|}
\hline
Model & Ensemble Weight (\%) & Public MAP & Private MAP  \\
\hline\hline
GatedNetVLAD(H1024C16) TTA[-1, 1]  & 0.22 & 0.7629 & 0.7556 \\ 
BERT(L2h12) TTA[-1, 1]             & 2.81 & 0.7729 & 0.7680 \\
BERTCrossMean(L2h8) TTA[-1, 1]     & 4.90 & 0.7748 & 0.7680 \\
BERT(L3h12) TTA[-1, 1]             & 13.88 & 0.7751 & 0.7688 \\
BERTCrossAttn(L2h8) TTA[-1, 1]     & 5.97 & 0.7758 & 0.7692 \\
BERTFirst(L2h12) TTA[-2, 1]        & 19.67 & 0.7792 & 0.7707 \\
MixNeXtVLAD(Iter 300) TTA[-1, 1]   & 20.03 & 0.7809 & 0.7711 \\
MixNeXtVLAD(Iter 60) TTA[0, 2]     & 17.26 & 0.7796 & 0.7721 \\
BERTMean(L2h12) TTA[-2, 1]         & 15.46 & 0.7802 & 0.7725 \\
Ensemble of all above             & 100.00 & 0.7944 & 0.7871 \\ 
\hline
\end{tabular}
\end{center}
\caption{Evaluation score of all our single models and ensembled model. For Gated NetVLAD model, H1024C16 means we use 1024 hidden size and 16 clusters. For MixNeXtVLAD model, iter 60 and 300 mean we use random sampling of 60 and 300 frames in input frame sequence, respectively. For BERT based models, L2h12 means we use 2 hidden layers and 12 heads. BERT without suffix means concatenating all the outputs from last layer of BERT. ``First'' suffix means using first token of BERT output layer. ``Mean'' suffix means averaging all the outputs from the last layer. ``Attn'' suffix means learnable weighted averaging of all the outputs from last layer of BERT. BERTCross is BERT with cross-modal learning. TTA means average ensemble of test-time augmentation with time shift. [a,b] means averaging the inferences on $\{i\in \mathbb{N}|a\le i\le b\}$ time shifted frames.}
\label{tab: evaluation}
\end{table*}

\section{Conclusion}
In this work, we present our approach in video segment classification leveraging video-level label data. We modify BERT model for video classification task and achieve competitive single model performance among our trained models. We ensemble BERT models with frame-level models such as Gated NetVLAD and NeXtVLAD according to the best weight learned by Bayesian optimization on local validation data. We propose simple yet effective test-time augmentation approach that further improves MAP@100K.

There are several future directions extending our work. 
\begin{enumerate}
    \item To achieve a better embedding representation of each frame based on other frames, we can pre-train BERT in the same fashion as in NLP by random masking of 15\% of frames. And we train the following models based on pre-trained one.
    \item We can apply video frame shift, dilute, enlarge as train time augmentation. Shift means randomly shifting left or right 1 time unit. Dilute means concatenating pre- and post- 2 frames and subsampling 5 frames out of 9. Enlarge means concatenating pre- and post- 2 frames.
    \item Given we have an augmentation approach, we can leverage large amount of unlabelled data by semi-supervised learning as described in Xie \etal~\cite{xie2019unsupervised}. Basically we want the KL divergence of predicted label probability between original and augmented data be as small as possible.
\end{enumerate}

{\small
\bibliographystyle{ieee_fullname}
\bibliography{TM_paper_bib}
}

% ----------------------------------------------------------------------------

\end{document}